\documentclass{article}

\usepackage{microtype}
\usepackage{graphicx}
\usepackage{subfigure}
\usepackage{booktabs} 
\usepackage{amsmath, amsthm, amssymb, amsfonts}
\usepackage{tcolorbox}
\usepackage{bm}

\usepackage{hyperref}

\usepackage[accepted]{icml2020}

\theoremstyle{definition}
\newtheorem{defn}{Definition}[section]

\icmltitlerunning{Abstraction Mechanisms Predict Generalization in Deep Neural Networks}

\begin{document}

\twocolumn[
\icmltitle{Abstraction Mechanisms Predict Generalization in Deep Neural Networks}




\begin{icmlauthorlist}
\icmlauthor{Alex Gain}{Alex Gain}
\icmlauthor{Hava Siegelmann}{Hava Siegelmann}
\end{icmlauthorlist}




]


\begin{abstract}
	
A longstanding problem for Deep Neural Networks (DNNs) is understanding their puzzling ability to generalize well. We approach this problem through the unconventional angle of \textit{cognitive abstraction mechanisms}, drawing inspiration from recent neuroscience work, allowing us to define the Cognitive Neural Activation metric (CNA) for DNNs, which is the correlation between information complexity (entropy) of given input and the concentration of higher activation values in deeper layers of the network. The CNA is highly predictive of generalization ability, outperforming norm-and-margin-based generalization metrics on an extensive evaluation of over 100 dataset-and-network-architecture combinations, especially in cases where additive noise is present and/or training labels are corrupted. These strong empirical results show the usefulness of CNA as a generalization metric, and encourage further research on the connection between information complexity and representations in the deeper layers of networks in order to better understand the generalization capabilities of DNNs.
			
\end{abstract}

\section{Introduction}

Deep neural networks (DNNs) have made big strides in recent years, improving substantially on state of the art results across many benchmarks and showing great generalization abilities \cite{lecun2015deep}. This is perhaps surprising given the large number of parameters, flexibility, and relative lack of explicit priors enforced in DNNs. Even more puzzling is their ability to memorize random datasets \cite{zhang2016understanding, yun2019small}. In part due to this, there has been a surge of work aimed at understanding the crucial factors to high-performing DNNs. Some studies have approached explaining generalization in DNNs via optimization arguments characterizing critical points \cite{dauphin2014identifying, kawaguchi2016deep, haeffele2017global}, the smoothness of loss surfaces \cite{nguyen2017loss, choromanska2015loss, li2017visualizing}, and implicit priors and regularization brought on by DNNs' learning methods \cite{mianjy2018implicit}, including that overparameterization itself leads to better learned optima and easier optimization \cite{arpit2019benefits, allen2019learning, oymak2019towards, allen2018convergence}. The work from \cite{morcos2018importance} provides insight by relating DNN generalization capability to reliance on single directions. Others have used information theory as a means of explaining the perfomance and inner-mechanisms of DNNs \cite{tishby2015deep, shwartz2017opening}. In \cite{lampinen2018analytic}, task structure of DNNs w.r.t. transfer learning is incorporated for tighter bounds on generalization error.

Complementary to these works, we approach the problem of understanding generalization of DNNs via the unconventional angle of applying the cognitive neuroscience concept of abstraction mechanisms \cite{calvo2008handbook, peters2017human, taylor2015global, shivhare2016cognitive, gilead2014mind} -- i.e. \textit{how representations are formed that compress information content while retaining only information which is relevant for generalization to unseen examples}. Specifically, we seek to analyze DNNs' representational patterns with comparison to abstraction mechanisms employed by the brain. Arguably, if we can quantify representational similarity (and dissimilarity) of these mechanisms in DNNs in comparison to the brain then it may aid in understanding their inner-mechanisms and could allow us to leverage current and ongoing neuroscience research. Additionally, it could bring new perspectives or understanding to the optimization, regularization, and information theory works cited in the previous paragraph. 

We focus on a study particularly amenable to algorithmic translation into conventional statistical learning settings (in our case, visual classification tasks) \cite{taylor2015global}, though we review similar works in the following related works section. As a summary, in the work from \cite{taylor2015global}, large-scale analyses of fMRI data were carried out to arrive at a precise notion of hierarchical abstraction in the brain. Roughly stated, deeper neurons, where neuron depth is defined as the distance from primary sensory cortices, show higher activation values when abstract behaviors or tasks are being performed. The opposite is true for less abstract tasks. Another way of stating the result is that the Pearson correlation between the linear regressed slopes of the activation values ordered by depth and the abstractness of the corresponding tasks is approximately 1. This neuroscience result is expounded on in section 3, and in the supplement, for clarity.

We translate these results into the form of a measure applicable to DNNs, which we term the Cognitive Neural Activation metric (CNA), which is computationally tractable, can be applied to any network archecture and dataset, and is easy to implement. 

The organization of this paper is as follows:

In section 3, we provide background on neuroscience results and introduce the precise definition of the CNA.

In section 4.1 and 4.2, we empirically relate test error and the CNA. We show three main results: \begin{enumerate}
	\item The loss landscape of classification error overlaps nicely with the CNA.
	\item High entropy images return up to 16 times larger test error across training epochs.
	\item Test error significantly correlates with the CNA across a breadth of image datasets and network architectures.
\end{enumerate}

In section 4.3, we adapt the CNA to predict the \textit{difference} between training and test error (termed the \textit{generalization gap}), and empirically validate its efficacy on a breadth of architectures and datasets, comprising over 100 dataset-architecture combinations. The architectures include Multi-layer perceptrons (MLPs), VGG-18, ResNet-18, and ResNet-101. The datasets include ImageNet, CIFAR-10, CIFAR-100, MNIST, Fashion-MNIST, SVHN, and corrupted labels counterparts (i.e. the same datasets with varyling levels training labels shuffled). The CNA outperforms recent metrics derived from theoretical generalization error bounds, especially in non-standard settings, showing significantly more robustness to additive noise.

\textbf{Contributions Summary:} We approach generalization in DNNs from an unconventional angle, where we connect abstractness mechanisms in the human brain to generalization error in DNNs in statistical learning settings. We explicitly translate a neuroscience result into a computationally tractable, differentiable mathematical expression, termed the Cognitive Neural Activation metric (CNA), that is easily implementable and can be applied to any network architecture. The CNA shows interesting connections to test error, and can be adapted such that it is predictive of generalization error in DNNs, outperforming recent generalization metrics based on theoretical generalization error bounds, and shows significantly more robustness to additive noise and label corruption. The strong emprical results encourage further work and exploration into this area of study.

\section{Related Work}

Comparisons of DNNs to cognitive neuroscience are often qualitative in nature or serve as loose analogies for illustration purposes only, e.g. historically much of the design of MLPs and CNNs were loosely motivated by computational neuroscience. Rigorous or empirical analyses are less common, however important progress has been made in characterizing representational similarity between DNNs and the brain. Seminal works \cite{yamins2013hierarchical, yamins2014performance} show significant similarity between the firing patterns of the visual cortex of primates and supervised models, as does the seminal work from \cite{khaligh2014deep} though with the caveat that unsupervised models do not. 

Recently, the Brain-Score has been developed \cite{schrimpf2018brain, kubilius2019brain} which uses the mean-score of neural predictivity (how well a given DNN predicts a single neuron's response in various visual systems of the primate) and behavioral similarity (how similar rates of correct and incorrect responses for specific inputs are between a given DNN and the primate) to rank the top-1 performance of state-of-the-art networks on ImageNet. The Brain-Score achieved significant correlation with top-performing networks' performance on ImageNet, showing neural representational similarity between the brain and DNNs can have useful predictive properties and, additionally, developed a high-performing shallow RNN based on it. 

The CNA mainly differs from the Brain-Score in that the CNA's primary application is in understanding and measuring the generalization gap between train and test sets across many different datasets and tasks, as opposed to ranking top-performing networks on ImageNet. Additionally, it is a differentiable geometric property or equation that, though grounded in empirical neuroscience data, does not actually use neuroscience data in its computation, i.e. it is a more basic function of a DNN's activation distribution and training distribution, not requiring empirical neuroscience data during inference at training or test. Thus, it is closer in nature to statistical-learning-based bounds on the generalization gap \cite{neyshabur2017exploring} and information-theoretic loss functions, e.g. the mutual information objective from Deep InfoMax \cite{hjelm2018learning}. 

Other related works utilizing cognitive neural activity for practical application include \cite{shen2019deep}, which used DNNs to reconstruct accurate, realistic-looking images from fMRI data, \cite{xu2018deeper}, which made use of empirical neuronal firing data to develop methods for explainable interpretability of DNNs, and \cite{arend2018single}, which show that many one-to-one mappings exist between individual neurons in DNNs and individual neurons in the brain as well as correspondences between population-level groups. In \cite{saxe2019mathematical}, they study the learning dynamics of linear networks and show that the learned representations share phemenoma observed in human semantic development. Lastly, in \cite{richards2019deep}, they give strong arguments for shifting the research paradigm of computational neuroscience towards utilizing three essential design components of DNNs: The objective functions, the learning rules, and the architectures.

Besides cognitive neuroscience work, the CNA primarily acts as a stand-in for margin-and-norm-based generalization metrics. We cover this in more detail in our main empirical results section.

\section{The Cognitive Neural Activation Metric}

	Due to its amenability to algorithmic implementation in statistical learning settings, we focus on a specific computational neuroscience work from \cite{taylor2015global}. Briefly stated, the study aimed to validate and characterize the extent to which and in what ways hierarchical representation occurs in the human brain. 
	
\begin{figure*}
			\centering
			\includegraphics[width=0.65\linewidth]{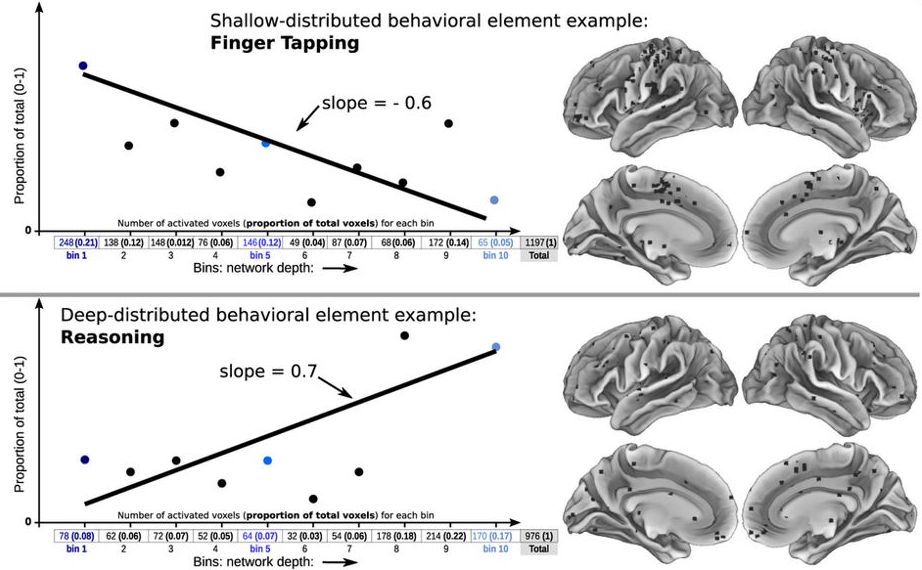}
			
			\textbf{(A)}
			\vspace{0.4cm}
			
			\makebox[\linewidth]{
				\fbox{\includegraphics[width=1.0\linewidth]{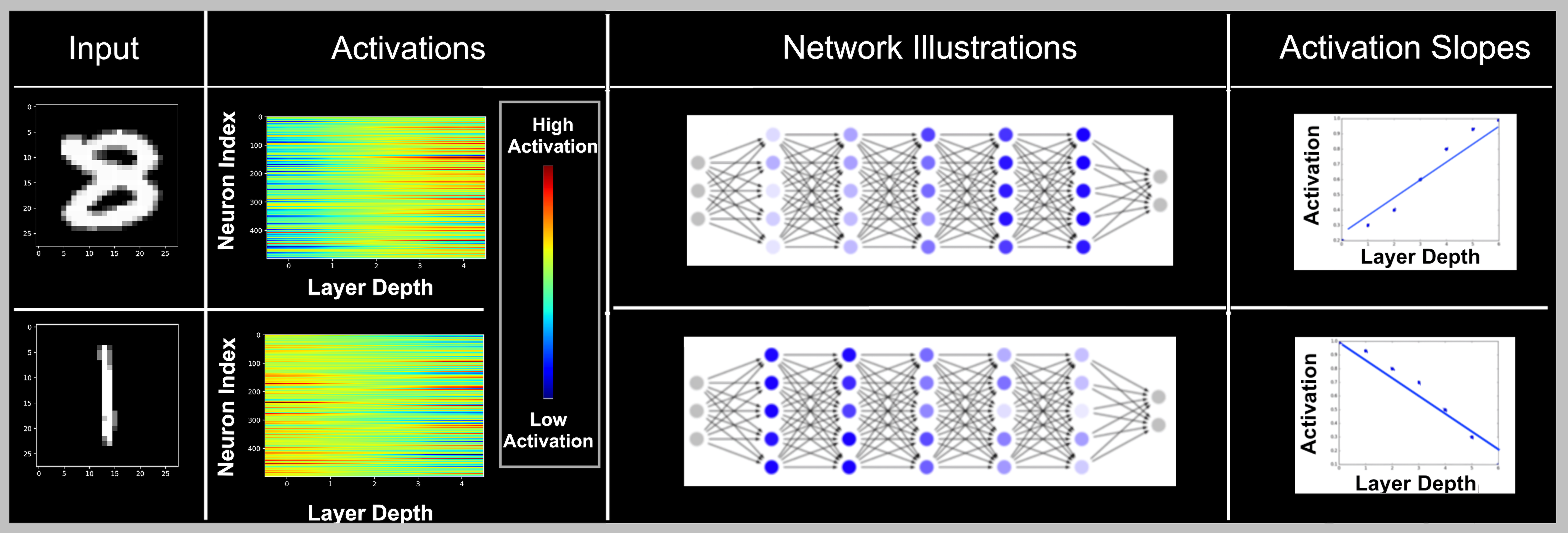}}
			}
			
			\vspace{0.4cm}
			\textbf{(B)}
			\vspace{0.1cm}
			
			\caption{\textbf{Illustrations of the Cognitive Neural Activation metric (CNA).} The two subfigures show correlations between data entropy and neuronal activation, with higher complexity input corresponding to higher values of activations in the deeper layers. This provides an intuitive visual of the CNA in a DNN (below) and the corresponding neuroscience principle (above). \textbf{(A)} Shows firing patterns in the human brain. X-axis is the bin number, with the first bin closest to the brain's inputs and 10th bin farthest away. Y-axis is the total activation per bin, normalized over numerous experiments of the same behaviors: tapping (top) and reasoning (bottom). The geometric slope of neuronal firing correlates with the behavior's entropy (figure from [12]). \textbf{(B)} Plot of neuronal activations and slope of a DNN trained on MNIST. The digit 8 has higher entropy and digit 1 has lower entropy, showing higher activations for the 8 digit in the deeper layers, as seen in the ``Activations'' column, where the  x-axis corresponds to the 5 hidden layers ordered by depth, and the y-axis plots 500 neuron values per layer. The ``network Illustrations'' shows sketches of DNN activity (only 5 layers shown) with the DNN showing a positive slope when processing the digit 8 and negative slope for the digit 1. Best viewed in color.}
			\label{fig:cna-illus}
		\end{figure*}

	They approached this by first hypothesizing that concrete functions occur ``earlier'' in the brain (closer to sensory input) whereas more abstract functions occur deeper in the brain (further from sensory input). Their main results can be summarized as: The abstraction level of tasks highly correlates with higher concentration of neural activity ``deeper'' in the brain. The computationally tractable, DNN analogue to this can be expressed as the correlation between information complexity of images and the concentration of high activation values in the deeper layers of the DNN. To quantify concentration of high activation values, we use the linearly regressed slope of the sum of activations by layer depth as a coarse measure\footnote{Of course, the relationship between neural activation values and depth is almost certainly nonlinear -- however, the purpose of the slope is to serve as a rough measure of neuronal activity, not to model the relationship between activation and depth.}, illustrated in Figure \ref{fig:cna-illus} for intuitive visualization of the principles of CNA.
	
	A high level definition and overview of the CNA can be seen in Figure \ref{fig:cna-equations}.
	
		\begin{figure*}
			\begin{tcolorbox}[boxsep=3mm]
				\begin{center}\textbf{Defining the CNA}\end{center}
				For a network architecture $A$ and dataset $X$ with $n$ data points, define
				\begin{enumerate}
					\item $\alpha{(x)}$ -- the information complexity (computed via histogram-binning approximation of Shannon entropy) of every datapoint $x \in X$, 
					\item $ \beta{(x)}$ -- the slope of neuronal activity of network $A$ when presented with $x \in X$,
					\item $\bm{\alpha}, \bm{\beta}$ -- the vectors of length $n$ comprising the complexity and slope values on the whole dataset $X$. 
				\end{enumerate}
				
				The CNA is defined by the Pearson correlation between the information complexity and the slope: 	\begin{equation}\rho_{\bf{\alpha,\beta}} = \frac{\text{cov}(\bm{\alpha},\boldsymbol{\beta})}{\sigma_\alpha \sigma_\beta}\end{equation}
				where $\text{cov}(\bm{\alpha},\boldsymbol{\beta})$ is the sample covariance of the two vectors: $\frac{1}{n-1}\sum_i(\boldsymbol{\alpha}_i - \overline{\boldsymbol{\alpha}})(\boldsymbol{\beta}_i - \overline{\boldsymbol{\beta}})$, $\overline{\boldsymbol{\alpha}}$ and $\overline{\boldsymbol{\beta}}$ are the means, and $\sigma_\alpha$,  and $\sigma_\beta$ are the sample standard deviations $\frac{1}{n-1}\sum_i(\boldsymbol{\alpha}_i - \overline{\boldsymbol{\alpha}})^2$ and $\frac{1}{n-1}\sum_i(\boldsymbol{\beta}_i - \overline{\boldsymbol{\beta}})^2$. 
				
			\end{tcolorbox}
			\caption{\textbf{CNA}: A high-level overview of the expressions comprising the CNA.}
			\label{fig:cna-equations}
		\end{figure*}

	We now proceed with a more rigorous, motivated treatment of the CNA and its neuroscience motivations.
	
	\subsection{Neuroscience Motivations}
	
	The work from \cite{taylor2015global} aimed to validate and characterize the extent to which and in what ways hierarchical representation occurs in the human brain. They approached this by first hypothesizing that concrete functions occur ``earlier'' in the brain (closer to sensory input) whereas more abstract functions occur deeper in the brain (further from sensory input). To assess this, they associated every region of interest in the brain with a number corresponding to its integrated distance from sensory cortices based on rsfMRI and DTI data, which they termed \textit{network depth}. Neurons were then binned according to the defined depth measure and activations for each bin were assessed via large-scale behavioral fMRI data analysis. 
	
	By ordering the bins by depth and then performing a linear regression on the aggregated activation values, a scalar-value slope can be obtained for a given behavior or task. Interestingly, these slopes were highly correlated with the abstraction of the task as obtained from Amazon Mechanical Turk (MTurk) surveys. This high correlation result is what we will focus on translating into a form that can be applied to DNNs. For a more detailed description of the neuroscience study, please refer to the supplement.
	
	We now precisely define the various components that comprise this result. Define the slope function $\beta$ which maps a task $T$ to its corresponding scalar-value slope as	
	\begin{equation*}
	\beta : \text{ task } T \longrightarrow \mathbb{R}
	\end{equation*}		
	Similarly, denote the abstractness measure function $\alpha$ which maps a task $T$ to its estimated measure of abstraction (estimated by MTurk in \cite{taylor2015global}) as	
	\begin{equation*}\alpha : \text{ task } T \longrightarrow \mathbb{R}\end{equation*}
	For $n$ tasks, denoted $T_1, T_2, \dots, T_n$, denote $\boldsymbol{\beta}\in \mathbb{R}^n$ as the vector of slopes where $\boldsymbol{\beta}_i = \beta(T_i)$ and denote $\mathbf{\alpha}\in \mathbb{R}^n$ as the vector of abstraction measurements where $\mathbf{\alpha}_i = \alpha(T_i)$.
	
	We denote the Pearson correlation between $\mathbf{\alpha}$ and $\boldsymbol{\beta}$ for $n$ tasks as 
	\begin{equation*}\rho_{\alpha,\beta} = \frac{\text{cov}(\mathbf{\alpha},\boldsymbol{\beta})}{\sigma_\alpha \sigma_\beta}\end{equation*}
	
	where $\sigma_\alpha$ and $\sigma_\beta$ denote the standard deviations of $\mathbf{\alpha}$ and $\boldsymbol{\beta}$. Finally, the neuroscience results can be restated as being equivalent to
	\begin{equation*}\rho_{\alpha,\beta} \approx 1\end{equation*}	
	for MTurk abstractness measurements $\mathbf{\alpha}$ and the slopes $\boldsymbol{\beta}$ obtained from the corresponding cognitive behavioral tasks. Thus, if we can determine appropriate definitions of $\mathbf{\alpha}$ and $\boldsymbol{\beta}$ for DNNs, then we now have a closed-form measure $\rho_{\alpha,\beta}$ for the extent to which a DNN employs similar abstraction mechanisms, where the activation distribution patterns are more similar as $\rho_{\alpha,\beta}$ approaches 1, unrelated as $\rho_{\alpha,\beta}$ approaches 0, and show the opposite pattern to that of the brain as $\rho_{\alpha,\beta}$ approaches -1. An illustration of a network closer to the $\rho_{\alpha,\beta} \approx 1$ pattern is shown in Figure \ref{fig:cna-illus}B.
		
	\subsection{Defining the CNA for DNNs}
	
	Now, the question remains, what are appropriate definitions for $\mathbf{\alpha}$ and $\boldsymbol{\beta}$? Defining $\boldsymbol{\beta}$ is straight-forward: Simply order the layers by depth (already explicitly defined for DNNs), define a method for aggregation of layer-wise activation values (e.g. mean, sum, etc.), and perform a linear regression on those values to arrive at a slope value.
	
	Defining $\mathbf{\alpha}$ is less straight-forward since abstractness is not well defined in statistical learning settings. We could make use of MTurk surveys, as was done in \cite{taylor2015global}, however this is not desirable for many reasons, especially since DNNs often require very large numbers of datapoints in order to be trained. To arrive at a well-defined notion for the CNA that is computationally tractable, we depart from biological equivalence and focus on the notion of compressibility as an analogue to abstraction. In fact, the MTurk results from \cite{taylor2015global} rely on the following definition of abstractness, 	
	\begin{quotation}
		``A process of creating general concepts or representations ... often with the goal of compressing the information content ... and retaining only information which is relevant.''
		\end{quotation} 	
	which corresponds to \textit{information compressibility}. Shannon entropy, easily implemented and calculated through histogram binning of feature values, is a lower bound for minimum description length, or Kolmogorov complexity, \cite{grunwald2004shannon} thus conveniently connecting the notion of abstractness to statistical learning settings\footnote{Ideally, the notional equivalence of compressibility and abstractness could be empirically vetted via human psychological experiments. Nonetheless, this analogue is useful in statistical learning settings given that the CNA has great predictive power, as shown in later sections. We focus on the statistical learning field, leaving further human psychological experiments as future work outside the scope of this paper.}.
			
	We now precisely define the CNA for a given network, parameterization, and batch of datapoints:
	
	For a data point $x \in \mathbb{R}^{k}$ with $k$ features, denote the informational complexity of $x$ as the scalar $\alpha(x)$, where $\alpha(x)$ is computed via a histrogram-binning Shannon entropy approximation. We define the slope $\beta(x)$ of a given network and datapoint $x$. Then, we define the Cognitive Neural Activation metric $CNA(\mathbf{X})$ for a given network, dataset $\mathbf{X} \in \mathbb{R}^{N \times k}$ of size $N$.
	
	\begin{defn}[Slope $\beta(x)$ of Network]
		For a given feedforward network with $L$ layers and input vector $x$, let $z_\ell^k(x)$ be the pre-activation state of neuron $k$ in layer $\ell$ given input vector $x$, let $n_\ell$ be the number of neurons in layer $\ell$, and let $z_\ell(x) = \sum_{k=1}^{n_\ell} z_\ell^k(x)$, i.e. the sum of the pre-activation values in layer $\ell$. Let $\bm{z}(x)$ be the vector of length $L$ where $[\bm{z}(x)]_\ell = z_\ell(x)$ for $\ell = 1, \dots, L$. Peforming a linear regression via least squares on the points $\{ (\ell, z_\ell(x)) \mid \ell = 1,\dots,L \}$, we obtain the slope $\beta(x)$ of the network.
	\end{defn}
	
	\begin{defn}[Cognitive Neural Activation metric $CNA(\mathbf{X})$ of Network]
		For a given feedforward network with $L$ layers and data matrix $\mathbf{X} \in \mathbb{R}^{N \times r}$, where $N$ corresponds to the number of samples: Let $x_i$ denote the $i^{\text{th}}$ row of $\mathbf{X}$. Let $\bm{\beta}(X)$ be the vector of length $N$ where $[\bm{\beta}(\mathbf{X})]_i = \beta(x_i)$ for $i = 1, \dots, N$ and let $\bm{\alpha}(\mathbf{X})$ be the vector of length $N$ where $[\bm{\alpha}(\mathbf{X})]_i = \alpha(x_i)$ for $i = 1, \dots, N$. $CNA(\mathbf{X})$ is defined as $corr(\bm{\beta}(\mathbf{X}),\bm{\alpha}(\mathbf{X}))$, where $corr$ denotes Pearson correlation.
	\end{defn}
	
	Thus, we arrive at our definition for the CNA. This can be straightforwardly extended to any CNN variant where $X$ instead exists in $\mathbb{R}^{N\times c \times h \times w}$, where $c$ denotes number of channels, $h$ denotes height of input, and $w$ denotes width of input. The average activation $z_\ell(x)$ of intermdiate layer $\ell$ is just calculated as the average of all activations for that layer, i.e. flattening the intermediate representation will yield the same result for $z_\ell(x)$.
	
	\section{Experimental Results}
	
	In this section, we show empirical validation of the CNA as a useful measure for understanding training in DNNs and their generalization error. 
	
	In sections 4.1 and 4.2, we show through empirical experiments and visualization of the loss landscape that there is a close connection between information complexity, CNA, and proper training of DNNs. 
	
	In section 4.3, we adapt the CNA to be indicative of the generalization gap, the difference between training error and test error. We compare the CNA to other norm-and-margin based generalization gap metrics, showing it outperforms them, especially on datasets where there is additive noise and/or label corruption.
	
	\subsection{How Does CNA Vary During Training?}
	
	Intuitively, the CNA is a measure of how similar a given DNNs' abstraction mechanisms are to that of the human brain on a given dataset. If a DNN generalizes well and has ``learned'' high-level concepts, it is conceivable that its CNA value, then, would be high. On the other hand, if the mechanisms by which DNNs abstract show no similarity to that of the brain, then we would expect to see no relationship between performance on a task and the CNA value.
	
\begin{figure}
		\centering
		\includegraphics[width=0.75\linewidth]{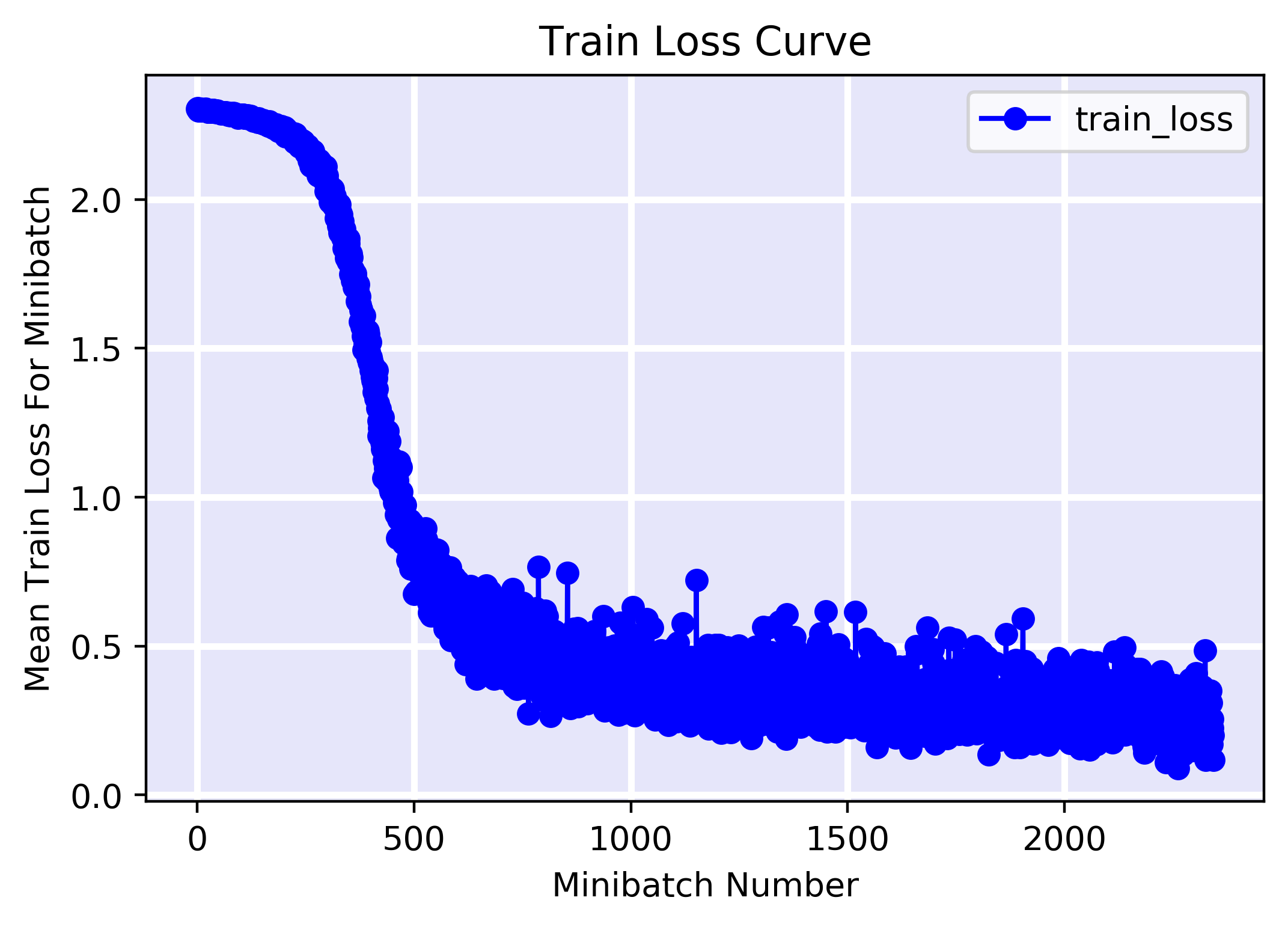}
		\vspace{0.1cm}
		\includegraphics[width=0.75\linewidth]{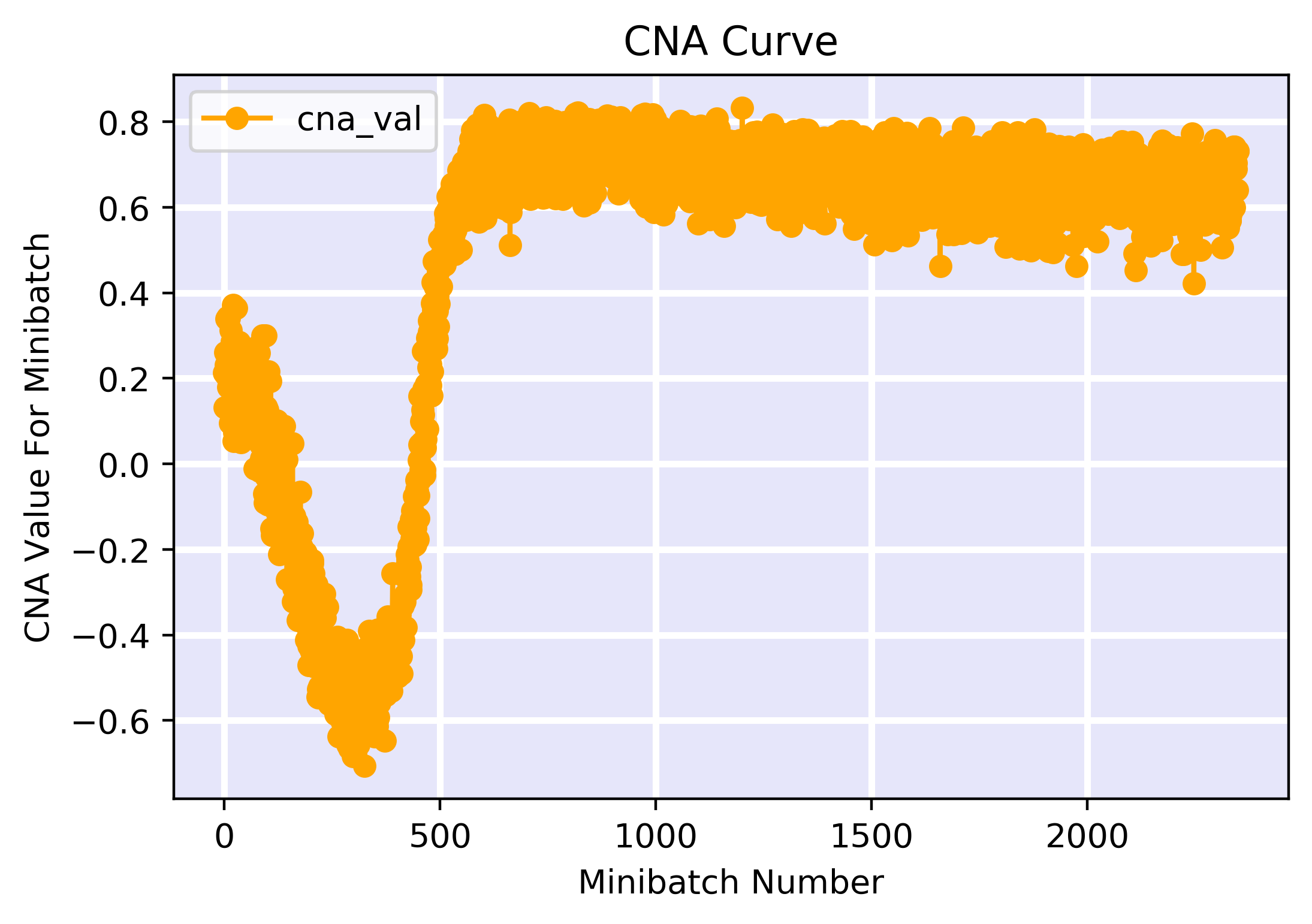}
		\caption{A simple MLP trained on MNIST with curves the training loss values (Top) and CNA values (Bottom) over training time. It is clear that the CNA shows a high correlation with training loss, with inflection points of both curves occuring at roughly the same timestep.}
		\label{fig:cnacurve}
		\end{figure}

	As a first step to investigating this, we train a simple MLP on the MNIST dataset and track the CNA value over training time. This is seen in Figure \ref{fig:cnacurve}. From the figure, the CNA clearly tracks training loss well, suggesting that the DNN is learning specific abstraction mechanisms over training time.	
	 Another interpretation of this result is simply that the loss landscape of the CNA and the supervised loss function (in this case, categorical cross entropy) are similar. If the gradient of the CNA function and the gradient of the supervised loss function are well-aligned throughout training time, then there is a significant chance for the CNA and the loss to be correlated. To investigate this, we record all neuronal activation values of the MLP at each minibatch update. We then perform PCA on the recorded neuronal activations values in order to visualize the optimization path of the network over training time, plotting the network state as a function of its principal components. Then, we sample from the x-y plane of principal component values and calculate the CNA value at each sample. This allows us to approximate and visualize the CNA loss landscape in the lower-dimensional space. The network converged to around 98\% test accuracy, showing that the high CNA value may be indicative of high test accuracy.
	 
\begin{figure}
		\centering
		\includegraphics[width=0.9\linewidth]{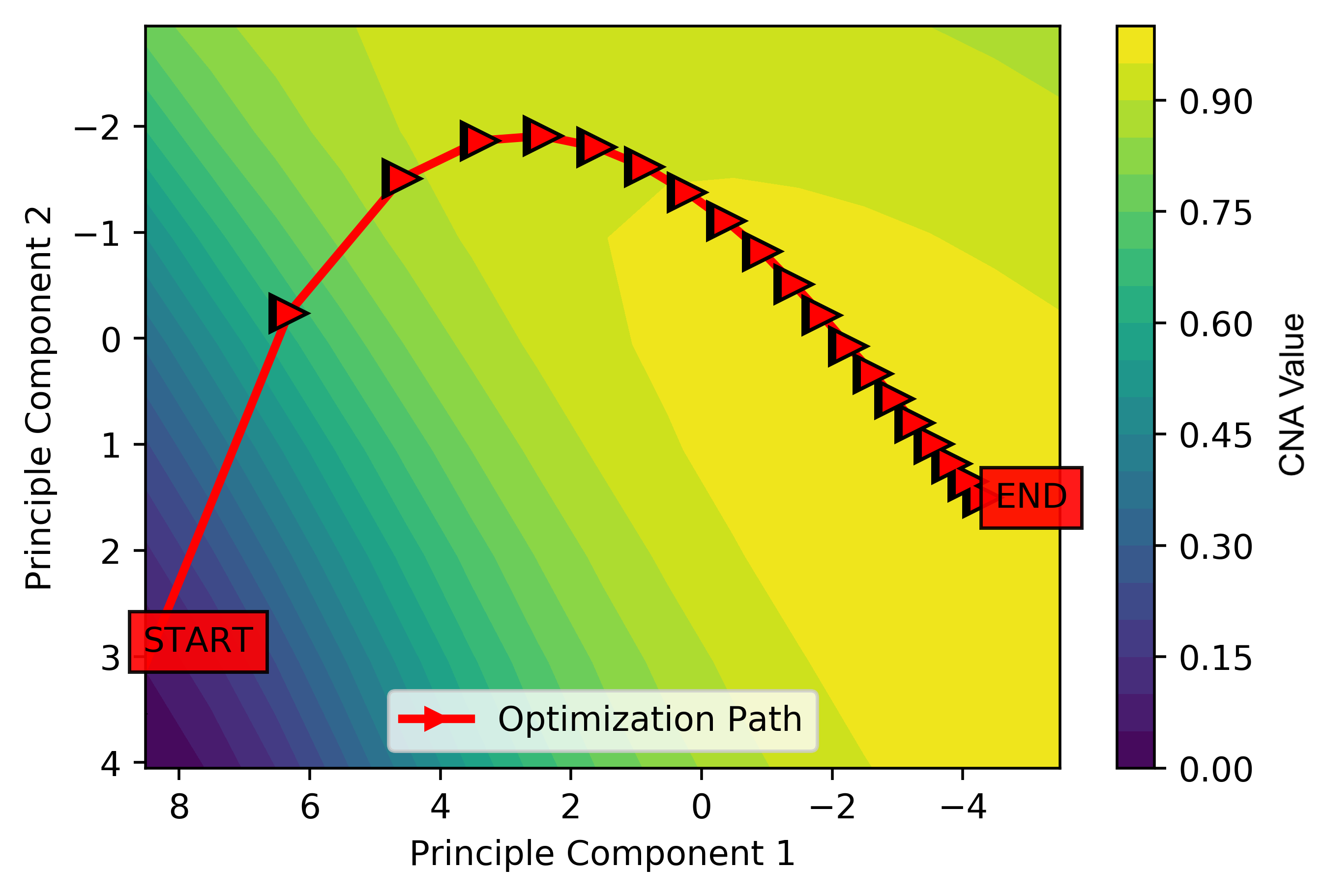}
		\caption{A low-dimensional visualization of the optimization path of an MLP over training time on MNIST, showing the network approximately traverses the CNA loss surface, despite being trained only on classification loss. The network state (all neuronal activation values) was recorded during each training step and then visualized using PCA (the red curve) with ``Start'' and ``End'' denoting the start and end points of training, with the x-axis and y-axis corresponding to the principal component values. The contour map behind the red curve (i.e. the CNA loss surface) was generated via sampling from the 2D principal component space and calculating the CNA value at each point. Best viewed in color.}
		\label{fig:optimization-path-2d-2}
		\end{figure}

	 These results and visualization can be seen in Figure \ref{fig:optimization-path-2d-2}.
		
		The results of Figure \ref{fig:optimization-path-2d-2} are perhaps surprising given that the CNA does not depend on labels, whereas the classification objective does. How, then, would the gradients of these two very different objectives be well aligned? A cursory look into the gradient expressions of the two terms gives a possible explanation, which we leave in the supplement in the interest of space. In short, the CNA and the supervised loss function could be well aligned in some cases where the error terms and information complexity terms for given datapoints significantly correlate.
		
\begin{figure}
			\centering
			\includegraphics[width=0.9\linewidth]{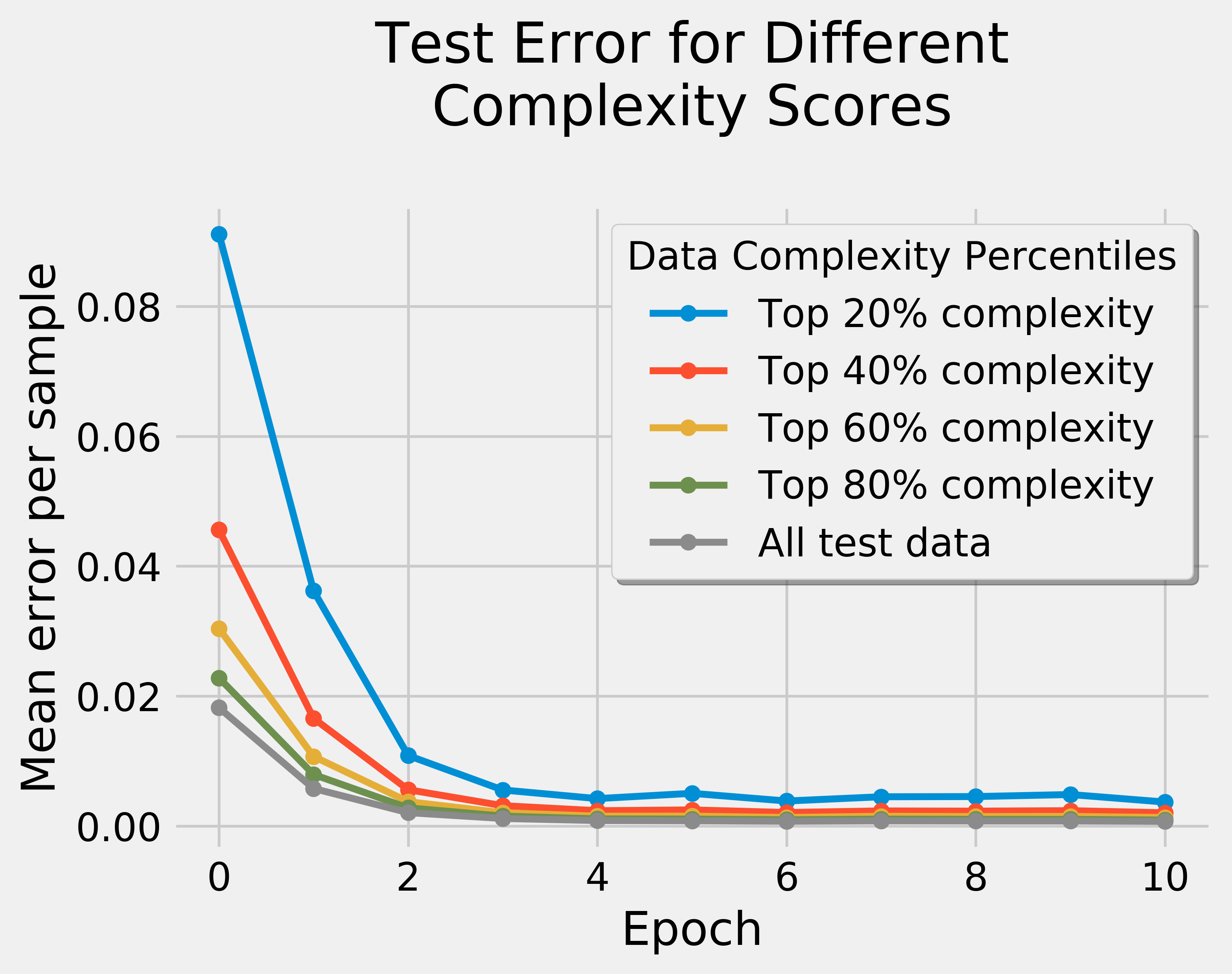}
			\caption{A plot of the mean test error across training time for bins of datapoints with varying levels of information complexity. There is a clear monotonic relationship between test error and complexity, especially at the beginning of training time when the CNA and test errors have the largest change. Best viewed in color.}
			\label{fig:error-complexity-curves}
			\end{figure}

		To give further credence to this explanation, we empirically analyze the relationship between the test error, which we denote as $\varepsilon$, and $\alpha$. We look at the mean error of the network on data points of varying $\alpha$ values to see if there is any relationship. Should the gradient argument hold true, we should expect that high-complexity datapoints will show larger error. This analysis is shown in Figure \ref{fig:error-complexity-curves}. We bin datapoints by their $\alpha$ values and plot their mean test error over training time, e.g. the blue curve corresponds to the datapoints with the top 20\% complexity values.

		Interestingly the figure shows a very clear, monotonic relationship between $\alpha$ and $\varepsilon$, especially in the beginning of training time. The differences between $\varepsilon$ at different complexity levels quickly decreases, though maintains its ordering, as training time increases. This makes sense given the network decreases its loss, and increases its CNA value, the most during the beginning epochs as was seen in Figure \ref{fig:cnacurve}.
		
		In summary, these results and figures show an interesting relationship between complexity of datapoints, CNA, training of DNNs, and test performance. These experiments by no means warrant broad conclusions -- more extensive analysis would be needed to draw more certain conclusions. Nonetheless, these raise questions as to how close the relationship is between CNA and training in DNNs, and whether there is a tight causal relationship between abstraction mechanisms and training.
		
		\subsection{Extensive Evaluation of CNA and Test Performance}
		
\begin{figure}
			\centering
			\includegraphics[width=0.8\linewidth]{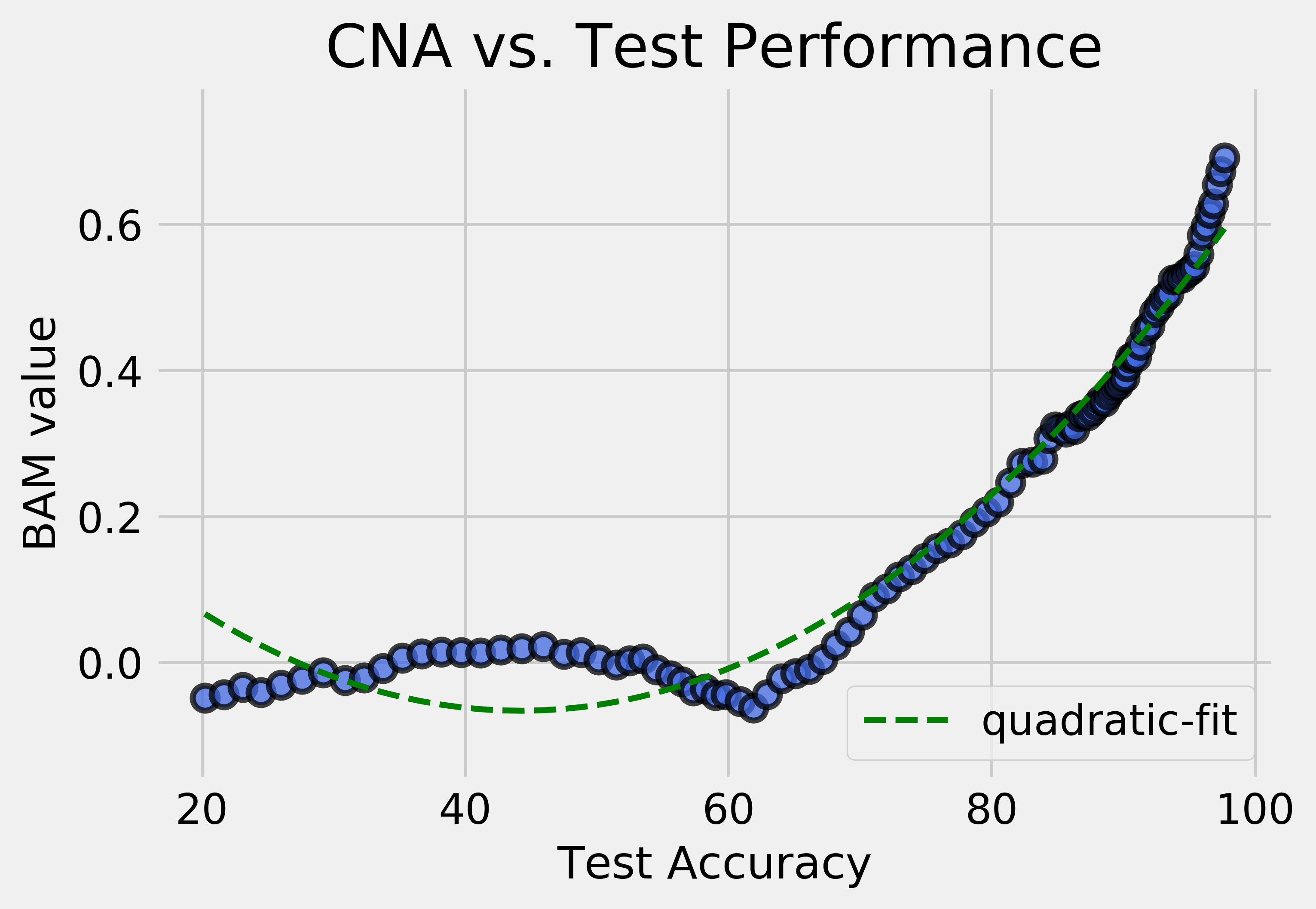}								
			\caption{The CNA strongly correlates with test accuracy. Here we show 147 dataset-architecture combinations (i.e., each dot represents a trained network) across six different datasets (ImageNet, CIFAR-10, CIFAR-100, SVHN, MNIST, Fashion-MNIST), four different architectures (MLP, VGG-18, ResNet-18, and ResNet-101), and measured at multiple stages of training (every 20 epochs). CNA correlates significantly with test accuracy, with a nearly linear relationship at greater than 70\%, suggesting neural activation properties of DNNs become more similar to the brain as classification results improve.}
			\label{fig:bam-gen-fig}
		\end{figure}

		We now carry out a far more extensive analysis of the CNA and test performance. We train four different architectures (MLP, VGG-18, ResNet-18, and ResNet-101) across six different datasets (MNIST, Fashion-MNIST, SVHN, CIFAR-10, CIFAR-100, and ImageNet) each, recording the network state, test error, and CNA value at every 20th epoch, comprising over 100 dataset-architecture combinations. These results are shown in Figure \ref{fig:bam-gen-fig}. There is a high correlation between CNA and test accuracy, suggesting close causal relationship between abstraction mechanisms and generalization ability. The result is especially convincing given the broad range of architectures and datasets tested on.

		This extensive evaluation gives more credence to the hypothesis that, as was seen in the previous subsection, the CNA, training in DNNs, and generalization in DNNs have an important relationship that warrants further study.
		
		\subsection{Extensive Evaluation of CNA and the Generalization Gap}
		
		Much work has been done on generalization bounds for DNNs based on norm and spectral properties of the weights \cite{neyshabur2017pac, neyshabur2017exploring}. Others include \cite{arora2018stronger}, which give bounds for DNNs based on compression properties, and \cite{neyshabur2018towards}, which give bounds based on overparameterization of DNNs. In \cite{jiang2018predicting}, a margin-based metric is developed and shows great success in correlating with the generalization gap, although with the drawback that a linear model needs to be fit between the generalization gap and margin parameters for each individual network and dataset. 
		
		As empirical validation of CNA, we focus on the very general setting where any network architecture, task, and dataset is allowed, and no model fitting with respect to the generalization gap is done, i.e. the developed metric must be predictive \textit{a priori}. 
	
\begin{figure}[h]	
			\centering
			
			\includegraphics[width=0.95\linewidth]{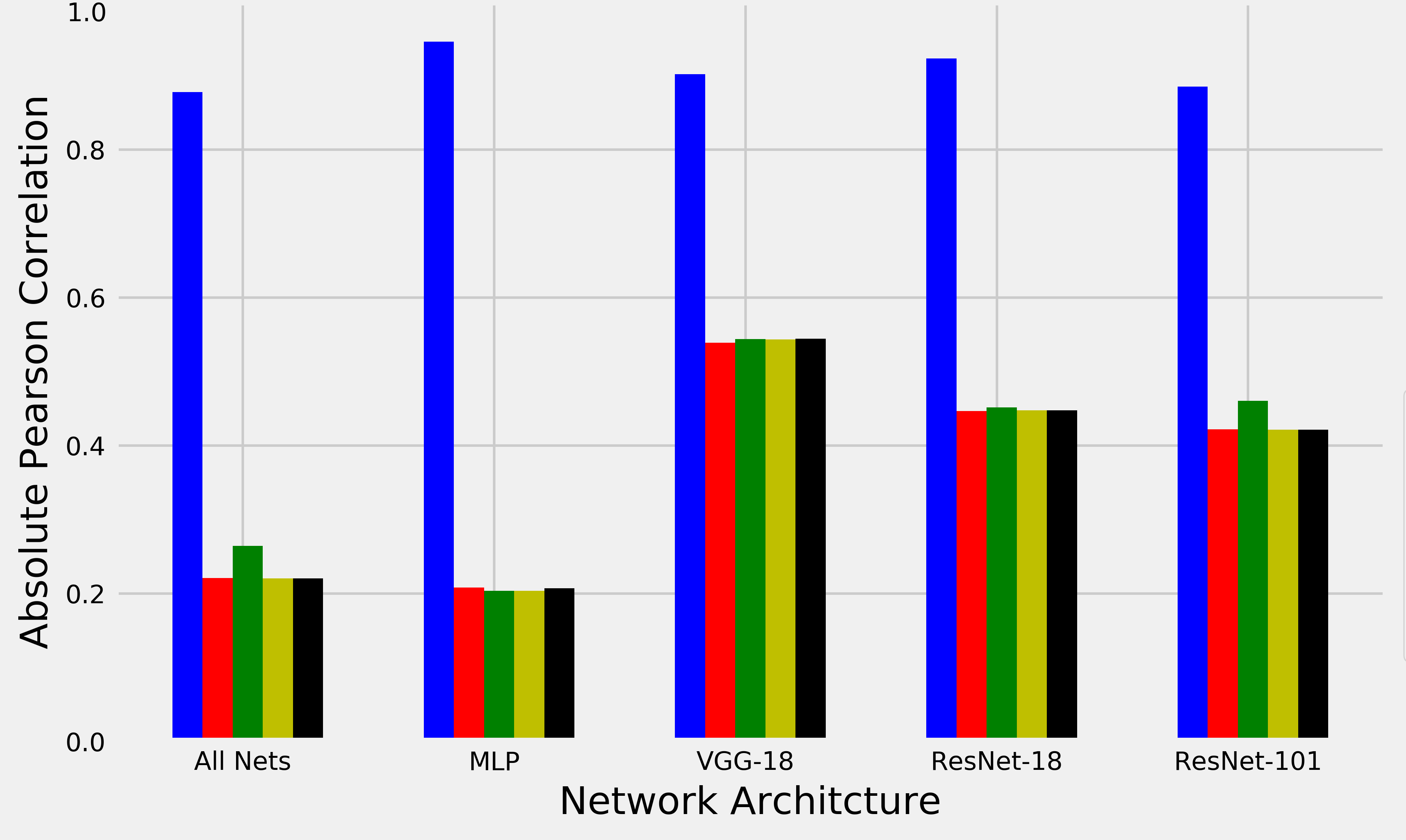}
			
			\vspace{0.1cm}
			\textbf{(A)}
			
			\includegraphics[width=0.95\linewidth]{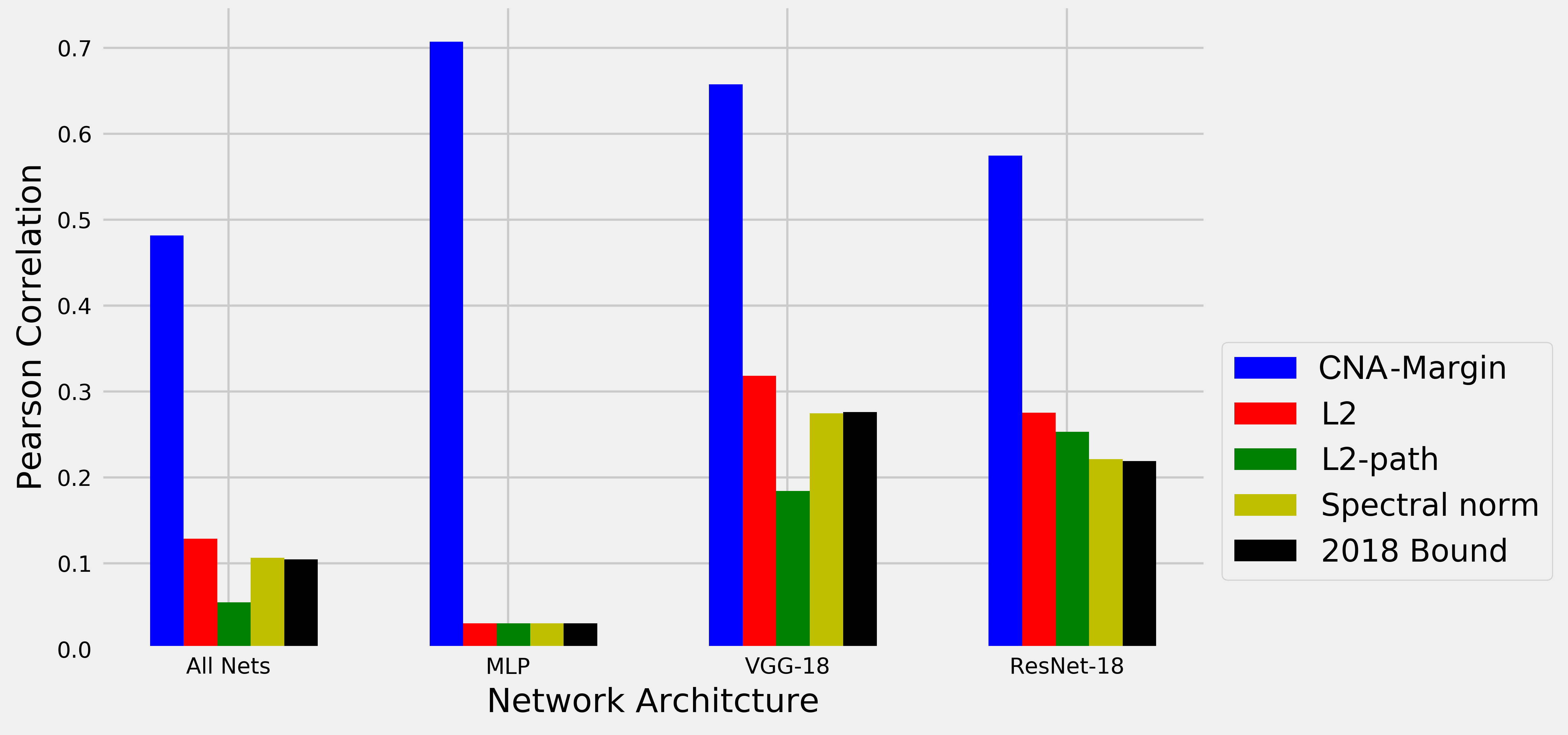}
			
			\vspace{0.1cm}
			\textbf{(B)}	
			
			\caption{We show the Pearson correlation of various generalization metrics with the train-test generalization gaps of over 177 combinations of networks. In total: Datasets (ImageNet, CIFAR-10, CIFAR-100, SVHN, MNIST, Fashion-MNIST) and network architectures (MLP, VGG-18, ResNet-18, ResNet-100), which were analyzed every 20 training epochs. We show the correlation conditional on network architecture as well as for all networks in aggregate (``All Nets'').  The CNA is comparatively robust to various types of corruption including in \textbf{(A)} where a random Gaussian noise dataset is included (detailed in the main text and supplement), and \textbf{(B)} where varying degrees of label corruption is present.}
			\label{fig:gen-bar-gap}
			
		\end{figure}

		To this end, we make use of the same 147 networks shown in Figure \ref{fig:bam-gen-fig}, and evaluate them on the CNA-Margin (a modified version of the CNA detailed in the supplement) and the competitive generalization metrics specified. Specifically, we record the generalization gap (the difference between train and test accuracy), each generalization metric for all networks, and calculate the Pearson correlation between each metric and the generalization gap. We additionally include a a Gaussian noise dataset. Each point in this dataset is drawn from the standard normal distribution of shape $3 \times 32 \times 32$ and labeled randomly to one of 10 classes -- the networks then memorize the training set. The norm-based metrics perform very poorly on this Gaussian noise dataset, whereas the CNA-Margin remains comparatively robust. These results are shown in In Figure \ref{fig:gen-bar-gap}A. The correlation is shown for all metrics, for each architecture, and for all architectures in aggregate (denoted ``All Nets'').
		
		Lastly, we train a subset of the networks on the same datasets, except with a varying degree of shuffled labels, ranging from 10\% to 50\% labels shuffled during training time. Similar to Figure \ref{fig:gen-bar-gap}A, the CNA-Margin remains robust compared to other metrics. All training details are included in the supplement, along with the chart of Figure \ref{fig:gen-bar-gap}A without the Gaussian dataset included for comparison purposes.
		
		\section{Conclusion}
		
		We provide principled motivation for a generalization metric inspired by cognitive neuroscience results. Interestingly, and perhaps suprisingly, the CNA shows connections with and predictive power for task performance in DNNs across a wide range of scenarios. Our CNA formulations show a practical use-case in predicting the generalization gap, outperforming margin-and-norm-based metrics, especially in the presence of dataset corruption. To our knowledge, our results comprise the most extensive study of generalization gap metrics in terms of breadth of dataset-architecture combinations considered. Through both small-scale and large-scale experiments, we show strong empirical support for the value of future work on understanding the relationship between abstract mechanisms, information complexity, and generalization capabilities in DNNs. 				

\bibliography{biblio}
\bibliographystyle{abbrvnat}

\end{document}